# Extraction of Projection Profile, Run-Histogram and Entropy Features Straight from Run-Length Compressed Text-Documents

Mohammed Javed[#1], P. Nagabhushan[#2], B.B. Chaudhuri[*3]

[#]*Department of Studies in Computer Science*
*University of Mysore, Mysore-570006, India*
[1]javedsolutions@gmail.com
[2]pnagabhushan@hotmail.com
[*]*Computer Vision and Pattern Recognition Unit*
*Indian Statistical Institute, Kolkata-700108, India*
[3]bbc@isical.ac.in

*Abstract-* Document Image Analysis, like any Digital Image Analysis requires identification and extraction of proper features, which are generally extracted from uncompressed images, though in reality images are made available in compressed form for the reasons such as transmission and storage efficiency. However, this implies that the compressed image should be decompressed, which indents additional computing resources. This limitation induces the motivation to research in extracting features directly from the compressed image. In this research, we propose to extract essential features such as projection profile, run-histogram and entropy for text document analysis directly from run-length compressed text-documents. The experimentation illustrates that features are extracted directly from the compressed image without going through the stage of decompression, because of which the computing time is reduced. The feature values so extracted are exactly identical to those extracted from uncompressed images.

## I. Introduction

Feature extraction is considered to be the most critical stage and plays a major role in the success of all image processing and pattern recognition systems [1]. Accordingly many sophisticated feature extraction techniques have been developed in the literature of document image analysis to deal with documents [2]. However, these techniques require that documents should be in uncompressed form. In real life, most of the document processing systems like fax machines [3], xerox machines, and digital libraries use compressed form to provide better transmission and storage efficiency. But the existing system has to decompress the document and process further. Thus decompression has become an unavoidable prerequisite which indents extra computation time and buffer space. Therefore, it would be appropriate to think of developing algorithms which can handle these compressed documents intelligently straight in their compressed formats.

Handling compressed documents directly for applications in image analysis and pattern recognition, is a challenging goal. The initial idea of working with compressed data was envisioned by a few researchers in early 1980's [4], [5]. Run-Length Encoding (RLE), a simple compression method was first used for coding pictures [6] and television signals [7]. There are several efforts in the direction of directly operating on document images in compressed domain. Operations like image rotation [8], connected component extraction [9], skew detection [10], page layout analysis [11] are reported in the literature related to run-length information processing. There are also some initiatives in finding document similarity [12], equivalence [13] and retrieval [14]. One of the recent work using run-length information is to perform morphological related operations [15]. In most of these works, they use either run-length information from the uncompressed image or do some partial decoding to perform the operations. However, to our best knowledge a detailed study on compressed documents from the viewpoint of computational cost, efficiency and validation with large dataset has not been attempted in the literature. In this research work, a novel idea of feature extraction straight from the compressed data is explored on binary documents as a case study to show the extraction of identical results from decompressed and Run-length Compressed Domains (RCD). The scope and overall architecture of RCD is shown in Fig-1.

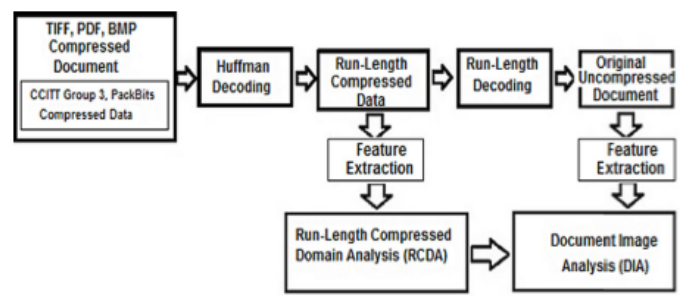

**Fig. 1: Proposed architecture for feature extraction and document analysis in uncompressed and Run-length Compressed Domains (RCD)**

Recently, there are few research efforts in direct processing of color images using JPEG and Wavelet compressed domain like face recognition [16], image indexing and retrieval [17]. Because of lossy compression techniques used with these color images, we cannot expect to extract identical features from their compressed and decompressed versions. However, an approximate to the original feature could be extracted from their compressed images, but we do not discuss these issues in this paper.

The objective of this research is not to define a new set of features to be extracted from compressed document images, but to extract the conventional features which are usually extracted from uncompressed documents, the novelty being to extract them straight from compressed version without spending on decompressing the image. As a consequence of this research, many efficient applications like text segmentation [18], document equivalence [19], word spotting in compressed documents are expected to be developed. Rest of the paper is organized as follows: section-2 describes some background details and proposed methods, section-3 discusses the experimental results and section-4 concludes the paper with a brief summary.

## II. PROPOSED METHODOLOGY

In this section, we will discuss the methods evolved by us, to extract the features straight from the compressed documents. Initially we give some background information necessary to understand our choice for working with binary documents popularly available in the form of TIFF image format. Then, we discuss about the data structure of run-length compressed data, which is then followed by brief idea about the projection profile, run-histogram and entropy.

### A. Background

There are many document image compression file formats available in the literature: BMP, PDF, JPEG and TIFF are commonly preferred formats. TIFF is very popular in handling monochrome or fax and handwritten documents and also preferred for archiving in digital libraries and in many network applications like printers, fax and xerox machines.

TIFF provides many built in compression algorithms: CCITT Group 3 (T.4)[20], CCITT Group 4 (T.6)[21] and many more. T.6 provides better compression ratio than T.4 and hence it was developed for archival purpose. However, the encoding process in T.6 is 2D and takes previous encoded line to encode the current line. In case of any single error during transmission, the whole document becomes unusable after decoding. Therefore, the natural choice for fax or network applications is T.4 standard, which facilitates both 1D and 2D coding. In this research work, we propose to use data from 1D (line by line) coding popularly known as Modified Huffman (MH) coding which is supported by all fax compression techniques. The basic and lossless compression scheme used here is Run-Length Encoding (RLE). In this paper, as we intend to show direct and identical feature extraction from compressed data, we chose to work on Run-length encoded binary documents.

In RLE, a run is a sequence of pixels having similar value and the number of such pixels is length of the run. The representation of the image pixels in the form of sequence of run values ($P_i$) and its lengths ($L_i$) is run-length encoding. It can be mathematically represented as ($P_i$, $L_i$). However, for binary images the coding is done with alternate length information of black and white pixels, which results in compact representation. The Table-I, gives the description of compressed data using RLE technique. The compressed data consists of alternate columns of number of runs of '0' and '1' identified as odd columns (1, 3, 5, ...) and even columns (2, 4, 6, ...) respectively. The proposed algorithms for extracting features like projection profile, run-histogram and entropy work on these type of data that will be frequently referred in this paper as compressed data.

TABLE I: Description of run-length compressed binary data

| Binary image Data | Compressed data | | | | | No. of Transitions (Positions) | |
|---|---|---|---|---|---|---|---|
| | 1 | 2 | 3 | 4 | 5 | +ve(0-1) | -ve(1-0) |
| 00000000000000 | 14 | 0 | 0 | 0 | 0 | 0 (0,0) | 0 (0,0) |
| 00110000111110 | 2 | 2 | 4 | 5 | 1 | 2 (3,9) | 2 (5,14) |
| 01111000111110 | 1 | 4 | 3 | 5 | 1 | 2 (2,9) | 2 (6,14) |
| 01111000111110 | 1 | 4 | 3 | 5 | 1 | 2 (2,9) | 2 (6,14) |
| 01111000111110 | 1 | 4 | 3 | 5 | 1 | 2 (2,9) | 2 (6,14) |
| 00110000000000 | 2 | 2 | 10 | 0 | 0 | 1 (3,0) | 1 (5,0) |
| 10000000000000 | 0 | 1 | 13 | 0 | 0 | 1 (1,0) | 1 (2,0) |
| 10000000000000 | 0 | 1 | 13 | 0 | 0 | 1 (1,0) | 1 (2,0) |
| 00100001111100 | 2 | 1 | 4 | 5 | 2 | 2 (3,8) | 2 (4,13) |
| 01111001111100 | 1 | 3 | 3 | 5 | 2 | 2 (2,8) | 2 (5,13) |
| 01111001111100 | 1 | 4 | 2 | 5 | 2 | 2 (2,8) | 2 (6,13) |
| 01111100000000 | 1 | 5 | 8 | 0 | 0 | 1 (2,0) | 1 (7,0) |
| 00000000000000 | 14 | 0 | 0 | 0 | 0 | 0 (0,0) | 0 (0,0) |

### B. Profiling

A projection profile is a histogram of the number of black pixel values accumulated along parallel lines taken through the document [22]. In the literature of document image analysis, projection profile methods have been used for skew estimation [23], text line segmentation [18], page layout segmentation [24] and in many more applications. For an uncompressed document of '$m$' rows and '$n$' columns, the mathematical representation for Vertical Projection Profile (VPP) [25] and Horizontal Projection Profile (HPP) are given below.

$$VPP(y) = \sum_{1 \leq x \leq m} f(x,y); HPP(x) = \sum_{1 \leq y \leq n} f(x,y)$$

However, the above formulae may not be directly applicable to work with our compressed data. We know that these data consists of alternate runs of black and white pixels which motivates us to add row-wise the alternate black-pixel-runs and skipping all white-pixel-runs to obtain VPP. The proposed method is simple and straight forward in getting VPP curve directly from the compressed data. However, obtaining HPP curve is a complex task. This is because the vertical information is not directly available in the compressed representation. So, we have developed an algorithm to compute HPP by pulling-out the run values from all the rows simultaneously using first two columns from compressed data shown in Table-I. In presence of zero run values in both the columns, the runs on the right are shifted to two positions leftwards. Thus for every pop operation, the run value is decremented by 1 and if the popped-out element is from first column then the transition value is 0 otherwise 1. This process is repeated for all the rows generating a sequence of column transitions from the compressed file which may be called *virtual decompression*. Thus addition of all these poppedout elements column-wise results in HPP curve. Consider a running example of a document ($m \times n$) and its compressed data of size $m \times n'$, where $n' < n$. The conventional VPP and HPP methods require $m \times n$ units of additions and memory space. Whereas, the time complexity of proposed methods are $m \times (n'/2)$ and $n \times m \times k$ respectively, where $k$ is the computing time for $n \times m$ virtual decompression (popping and shifting) and additions. The comparative analysis of these methods is demonstrated with experiments in section-3.

*C. Run-Histogram*

Generally a histogram of a digital image represents the frequency of number of intensity values present in the image [26]. However for binary images, a simple histogram would not give much information needed for document analysis. This motivates us to define run-histogram on compressed data, which has been recently applied for document retrieval and classification [27] using uncompressed documents. The compressed data in Table-I contains alternate columns of white and black pixel runs. Using this data structure, we compute the frequency of each run separately for even and odd columns to obtain black-pixel and white-pixel run-histograms. Combining these two run-histograms we get black-white run-histogram. Their time complexities are $m \times (n'/2)$, $m \times (n'/2)$ and $m \times n'$ respectively.

The above run-histogram can be used to analyze the amount of information and blank space available within the document. Consider a simple application; a run-histogram along the peaks of projection profile shown in Fig-2c can give an approximate idea of number of words and their separation information existing within the compressed text line. Very recently the work of [27] incited us to extend the idea of run-histogram to logarithmic scale run-histogram. The quantized length of the runs in a logarithmic scale is as follows:[1], [2], [3 −4], [5 − 8], [9 − 16], ..., [129−]. The number of bins required can be created based on the application. The special feature of these type of run-histograms is that the runs are clustered based on the logarithmic scale. This may find real time applications in many areas. One such application could be counting number of blank lines (run value equal to column size) within the compressed document. Thus, we hope that proposed techniques will be useful for many other applications of document processing on compressed data.

*D. Entropy*

In this research work, we demonstrate the extraction of Conventional Entropy Quantifier (CEQ) and Sequential Entropy Quantifier (SEQ) proposed by [19], straight from the compressed data. The details regarding the idea, motivation and formulation of CEQ and SEQ can be obtained in [19]. CEQ measures the energy contribution of each row by considering the probable occurrence of +ve (0-1) and -ve (1-0) transitions among the total number of pixels present in that particular row. On the other hand SEQ measures the entropy at the position of occurrence of these transitions. However from the compressed data, these information are easily available in the alternate even (+ve) and odd (-ve) columns as shown in Table-I. In each row, counting the number of even or odd columns of non-zero runs, leaving the first column (run) gives the number of +ve or -ve transitions respectively.

Whereas, the summation of all the previous runs incremented by 1 is the position of the transition point at a particular column as shown in Table-I. The time complexity for CEQ is $m \times (n' - 1)$ search and $m$ entropy computations, whereas for SEQ is $m \times (n' - 1)$ and $m \times (n' - 1)$.

The mathematical formulations for computing horizontal entropy using CEQ and SEQ from compressed data are given below.

$$\text{For CEQ, } E(t) = p * \log(\tfrac{1}{p}) + (1-p) * \log(\tfrac{1}{(1-p)})$$

where $t$ is the transition from $0 - 1$ and $1 - 0$, $E(t)$ is the entropy. Accordingly $p$ is ratio of number of +ve or –ve transitions to that total number of transitions possible in a row ($n'-1$, where $n'$ is summation of all runs in a row). Entropy at even and odd columns are computed independently and termed as $E^+(t)$ and $E^-(t)$ respectively and the total entropy $E(t)$ of each row is the summation of $E^+(t)$ and $E^-(t)$. For SEQ, the probable occurrence $p$ is replaced by the position parameter *pos*, which indicates the position of transition point. If the transition occurs between two columns $C_{\beta1}$ and $C_{\beta2}$ in row $r_\alpha$ then corresponding row entropy is formulated as:

$$E(\beta) = \tfrac{r_\alpha}{m}\left(\tfrac{pos}{n} * \log \tfrac{n}{pos} + (m - \tfrac{pos}{n}) * \log \tfrac{m}{m*n-pos}\right)$$

where $\beta = 1, ...m$. Total entropy for each row is the summation of entropy at even and odd columns represented as $E^+(\beta)$ and $E^-(\beta)$ respectively. The total horizontal entropy using CEQ and SEQ are the summation of $E(t)$ and $E(\beta)$ of all rows in the compressed data. In order to compute the vertical entropy from compressed data, the *virtual decompression* algorithm discussed earlier is applied to extract column transition values and entropy is computed using the formulas proposed by [19]. Due to page constraints, we do not discuss these details in this paper. The experimental results and computational efficiency of horizontal entropy using CEQ and SEQ is demonstrated in section-3.

III. EXPERIMENTAL RESULTS AND DISCUSSIONS

In this research work, we have proposed algorithms to extract features directly from run-length compressed documents. In real time, the run-length data has to be extracted from the compressed files using partial huffman decoding as shown in the Fig-1. However in order to validate our proposed ideas, we have considered ten noise free, manually run-length compressed machine printed text-documents for experimentation.

Consider a decompressed version of a sample document in Fig-2a; the horizontal and vertical projection profile curves and run-histogram feature obtained with our proposed methods using compressed and decompressed data of this document are given by Fig-2. The document in Fig-2a consists of eight text lines which are clearly visible by the number of peaks present in profile diagram extracted from its compressed (Fig-2c) and decompressed data (Fig-2e). Further processing of the this projection profile leads to text line segmentation. The experimental results in Fig-2 show that the features such as HPP, VPP and run-histogram extracted from compressed and decompressed data are identical.

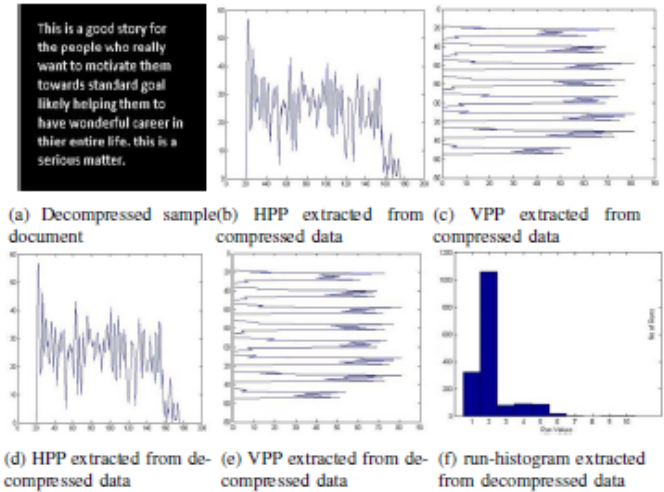

Fig. 2: Sample document and its corresponding horizontal and vertical projection profile curves, run-histogram extracted from its compressed (Fig-2b,Fig-2c and Fig-3a) and decompressed (Fig-2d,Fig-2e and Fig-4b) versions

Histograms are used to plot the frequency of the contents present in the image. However for compressed data, we have used histogram to plot the run frequency information. This results in three different run-histograms: black-pixel-run, whitepixel-run and combining both we get black-white-pixel-run. Fig-3a, Fig-3b and Fig-3c shows these three different types of run-histograms for the sample document considered above. Also as discussed in the earlier section, the three different run-histograms obtained from the compressed document using logarithmic scale are shown in Fig-3d, Fig-3e and Fig-3f.

The Table-II gives the total computation time saved in extracting different features proposed from 10 compressed documents. The total computation time saved is calculated by the formula $((T1-T2)/T1) \times 100$, where $T1$ and $T2$ are total computation time for extracting features with and without decompression time ($D$) respectively. Apart from saving computation time, the proposed feature extraction methods are themselves efficient as shown by the feature extraction time graphs in Fig-4a, Fig-4c, Fig-4g and Fig-4i. On the other hand we experience computation time loss in case of HPP due to virtual decompression as shown by Table-II, Fig-4e and Fig-4f. But the over all performance of feature extraction directly from compressed data is appealing.

We have used CEQ and SEQ entropy measures to compute horizontal entropy directly from the compressed document. The entropy values obtained match exactly with that of conventional method as shown in Table-III. The total computation time saved for a sample of 10 compressed and decompressed documents is tabulated in Table-II.

TABLE III: CEQ and SEQ feature comparison

| Document | CEQ | | SEQ | |
| --- | --- | --- | --- | --- |
| | Compressed | decompressed | Compressed | decompressed |
| Sample Document | 22.3466 | 22.3466 | -635170 | -635170 |

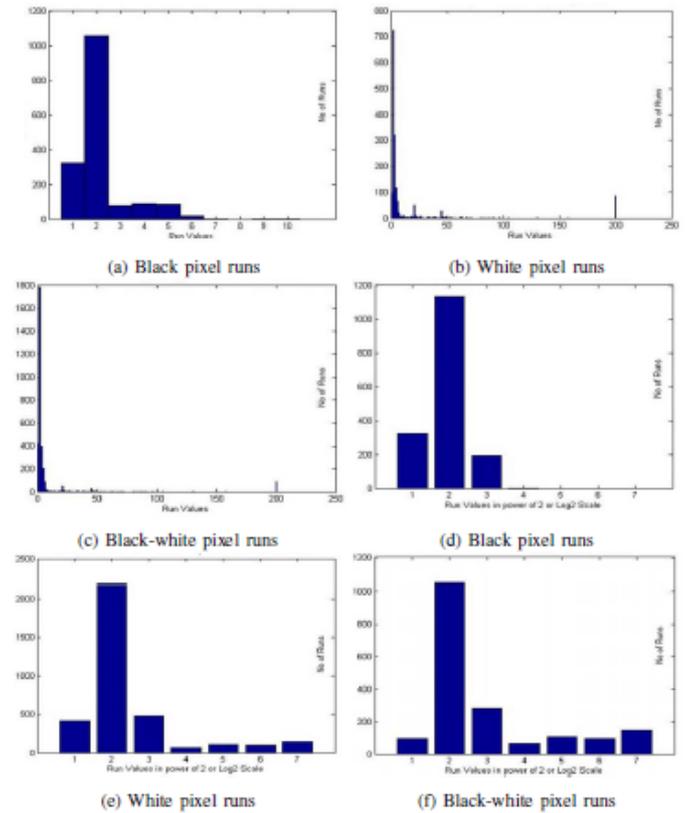

Fig. 3: Three different run-histograms resulting from run-length compressed document(Fig-3a,Fig-3b and Fig-3c) and three different run-histograms resulting from run-length compressed document using Logarithmic Scale(Fig-3d,Fig-3e and Fig-3f).

The idea of feature extraction directly from the compressed data could be extended to gray scale or color images. However, the possibility of getting the identical features from the compressed color images as demonstrated in this paper on binary documents, could be very less. The simple reason for this may be the usage of lossy techniques during compression. Nevertheless, we strongly feel that an approximate to the original features could be extracted from these images. However, these are unexplored and open research issues which can be taken as an extension work of this research paper.

IV. CONCLUSION

In this research work, a novel idea of feature extraction straight from the compressed documents is demonstrated. Apart from computational efficiency, the proposed methods avoid decompression which results in saving of considerable amount of computing resources. From the experiments, we also show that the features such as projection profile, runhistogram and entropy extracted from run-length compressed data are efficient and results are 100% accurate to that of existing methods over the decompressed data. As a result of this research, we are further motivated to think of extending this work to color images and also developing efficient applications like document equivalence, segmentation, word spotting, retrieval on compressed data.

TABLE II: Total feature extraction time (in *seconds*)

| Sample | Vertical Projection Profile | | | | Horizontal Projection Profile | | | | Run-Histogram | | | | CEQ | | | | SEQ | | | |
|---|---|---|---|---|---|---|---|---|---|---|---|---|---|---|---|---|---|---|---|---|
| | T2 | D | T1 | Time Saved(%) | T2 | D | T1 | Time Loss(%) | T2 | D | T1 | Time Saved(%) | T2 | D | T1 | Time Saved(%) | T2 | D | T1 | Time Saved(%) |
| 10 Documents | 0.043 | 1.80 | 2.012 | 97.86 | 5.417 | 1.80 | 2.233 | 58.46 | 3.50 | 1.80 | 13.11 | 73.30 | 0.112 | 1.80 | 2.217 | 94.60 | 0.165 | 1.80 | 2.367 | 93.05 |

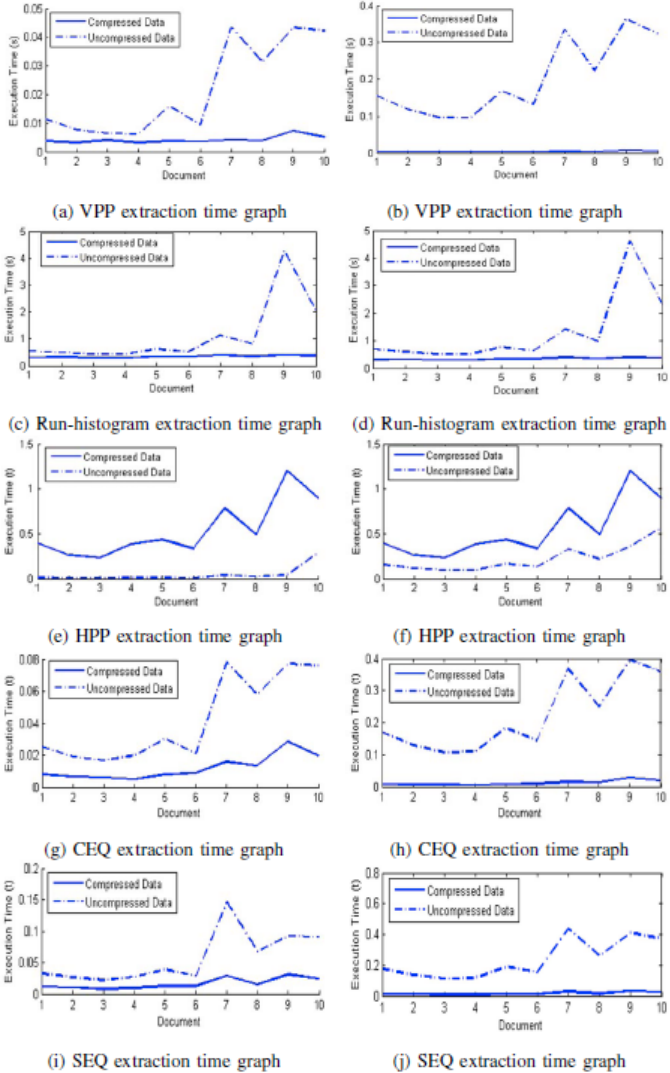

**Fig. 4: Feature extraction time graphs from compressed and uncompressed data with (Fig-4b, Fig-4d, Fig-4f, Fig-4h, Fig-4j) and without (Fig-4a, Fig-4c, Fig-4e, Fig-4g, Fig-4i) decompression time**

REFERENCES

[1] M. S. Nixon and A. S. Aguado, *Feature Extraction and Image Processing*. Newnes, 1 ed., 2002.
[2] K. Jung, K. I. Kim, and A. K. Jain, "Text information extraction in images and video: a survey," *Pattern Recognition*, vol. 37, no. 5, pp. 977–997, 2004.
[3] K. R. McConnell, D. Bodson, and R. Schaphorst, *FAX: Digital FacsimileTechnology and Applications*. Artech House, 1989.
[4] G. Grant and A. Reid, "An efficient algorithm for boundary tracing and feature extraction," *Computer Graphics and Image Processing*, vol. 17,pp. 225–237, November 1981.
[5] T. Tsuiki, T. Aoki, and S. Kino, "Image processing based on a runlength coding and its application to an intelligent facsimile," *Proc. Conf. Record, GLOBECOM '82*, pp. B6.5.1–B6.5.7, November 1982.
[6] J. Capon, "A probabilistic model for run-length coding of pictures," *IRE Transactions on Information Theory*, vol. 5, pp. 157–163, 1959.
[7] J. Limb and I. Sutherland, "Run-length coding of television signals,"*Proceedings of IEEE*, vol. 53, pp. 169–170, 1965.
[8] Y. Shima, S. Kashioka, and J. Higashino, "A high-speed algorithm for propagation-type labeling based on block sorting of runs in binary images," *Proceedings of 10th International Conference on Pattern Recognition (ICPR)*, vol. 1, pp. 655–658, 1990.
[9] E. Regentova, S. Latifi, S. Deng, and D. Yao, "An algorithm with reduced operations for connected components detection in itu-t group 3/4 coded images," *IEEE Transactions on Pattern Analysis and Machine Intelligence*, vol. 24, pp. 1039 – 1047, August 2002.
[10] A. L. Spitz, "Analysis of compressed document images for dominant skew, multiple skew, and logotype detection," *Computer vision and Image Understanding*, vol. 70, pp. 321–334, June 1998.
[11] E. Regentova, S. Latifi, D. Chen, K. Taghva, and D. Yao, "Document analysis by processing jbig-encoded images," *International Journal on Document Analysis and Recognition (IJDAR)*, vol. 7, pp. 260–272, 2005.
[12] D. S. Lee and J. J. Hull, "Detecting duplicates among symbolicallycompressed images in a large document database," *Pattern Recognition Letters*, vol. 22, pp. 545–550, 2001.
[13] J. J. Hull, "Document image similarity and equivalence det," *International Journal on Document Analysis and Recognition (IJDAR'98)*, vol. 1, pp. 37–42, 1998.
[14] Y. Lu and C. L. Tan, "Document retrieval from compressed images,"*Pattern Recognition*, vol. 36, pp. 987–996, 2003.
[15] T. M. Breuel, "Binary morphology and related operations on run-length representations," *International Conference on Computer Vision Theory and Applications - VISAPP*, pp. 159–166, 2008.
[16] M. S. Moin and A. S. Moghaddam, "Face recognition in jpeg compressed domain: a novel coefficient selection approach," *Signal, Image and Video Processing*, pp. 1–3, 2013.
[17] G. Schaefer and D. Edmundson, "Dc stream based jpeg compressed domain image retrieval," *Active Media Technology Lecture Notes in Computer Science*, vol. 7669, pp. 318–327, 2012.
[18] M. Arivazhagan, H. Srinivasan, and S. Srihari, "A statistical approach to line segmentation in handwritten documents," *Proc. SPIE 6500, Document Recognition and Retrieval XIV, 65000T*, January 2007.
[19] S. D. Gowda and P. Nagabhushan, "Entropy quantifiers useful for establishing equivalence between text document images," *International Conference on Computational Intelligence and Multimedia Applications*, pp. 420 – 425, 2007.
[20] CCITT-Recommedation(T.4), "Standardization of group 3 facsimile apparatus for document transmission, terminal equipments and protocols for telematic services, vol. vii, fascicle, vii.3, geneva," tech. rep., 1985.
[21] CCITT-Recommedation(T.6), "Standardization of group 4 facsimile apparatus for document transmission, terminal equipments and protocols for telematic services,vol. vii, fascicle, vii.3, geneva," tech. rep., 1985.
[22] R. Kasturi, L. O. Gorman, and V. Govindaraju, "Document image analysis: A primer," *Sadhana Part 1*, vol. 27, pp. 3–22, 2002.
[23] H. Baird, "Skew angle of printed documents," *Proceedings of SPSE's 40th Annual Conference and Symposium on Hybrid Imaging Systems*,pp. 21–24, 1987.
[24] G. Nagy, S. Seth, and M. Viswanathan, "A prototype document image analysis system for technical journals," *Computer*, vol. 25, no. 7, pp. 10–22, 1992.
[25] L. Likforman-Sulem, A. Zahour, and B. Taconet, "Text line segmentation of historical documents: a survey," *International Journal of Document Analysis and Recognition (IJDAR)*, vol. 9, pp. 123–138, April 2007.
[26] R. C. Gonzalez and R. E. Woods, *Digital Image Processing, Third Edition*. Pearson, 2009.
[27] A. Gordo, F. Perronnin, and E. Valveny, "Large-scale document image retrieval and classification with runlength histograms and binary embeddings," *Pattern Recognition*, vol. 46, pp. 1898–1905, July 2013.